\pdfoutput=1

\documentclass[11pt]{article}
\usepackage{authblk}
\usepackage[]{acl}

\usepackage{times}
\usepackage{latexsym}

\usepackage[T1]{fontenc}

\usepackage[utf8]{inputenc}
\usepackage{booktabs}

\usepackage{microtype}
\usepackage{amsmath}
\usepackage{graphicx}
\usepackage{subfigure}
\usepackage{amsfonts}
\usepackage{multirow}
\usepackage{listings}
\usepackage{url}
\usepackage[ruled,linesnumbered]{algorithm2e}  
\usepackage{CJKutf8}
\usepackage{graphicx}
\usepackage{subfigure}
\usepackage{supertabular}
\usepackage{subfigure}
\usepackage{dcolumn,booktabs}
\usepackage{tikz}
\usepackage[right]{rotating}

\usepackage{inconsolata}

\title{MoSECroT: Model Stitching with Static Word Embeddings \\ for Crosslingual Zero-shot Transfer}

\author[]{\bf Haotian Ye$^{\text *}$}
\author[]{\bf Yihong Liu$^{\text *}$}
\author[]{\bf Chunlan Ma$^{\text *}$}
\author[]{\bf Hinrich Sch\"utze}

\affil{Center for Information and Language Processing, LMU Munich \\ Munich Center for Machine Learning (MCML)
 \protect\\ \texttt{\{yehao, yihong, chunlan\}@cis.lmu.de}} 

\def\secref#1{\S\ref{sec:#1}}
\def\seclabel#1{\label{sec:#1}}

\begin{document}
\maketitle
\def\thefootnote{*}\footnotetext{Equal contribution.}\def\thefootnote{\arabic{footnote}}
\begin{abstract}
Transformer-based pre-trained language models (PLMs) have
achieved remarkable performance in various natural language
processing (NLP) tasks. However, pre-training such models
can take considerable resources that are almost only
available to high-resource languages. On the contrary,
static word embeddings are easier to train in terms of
computing resources and the amount of data required. In this paper,
we introduce \textbf{MoSECroT}
(\textbf{Mo}del \textbf{S}titching with Static
Word \textbf{E}mbeddings for \textbf{Cro}sslingual
Zero-shot \textbf{T}ransfer), a novel and challenging task
that is especially relevant to low-resource languages for
which static word embeddings are available.  To tackle the
task, we present the first framework that leverages relative
representations to construct a common space for the
embeddings of a source language PLM and the static word
embeddings of a target language. In this way, we can train
the PLM on source-language training data and perform
zero-shot transfer to the target language by simply swapping
the embedding layer.  However, through extensive experiments
on two classification datasets, we show that although our
proposed framework is competitive with weak baselines when
addressing MoSECroT, it fails to achieve competitive results
compared with some strong baselines.  In this paper, we
attempt to explain this negative result and provide several
thoughts on possible improvement.
\end{abstract}

\section{Introduction}
The emergence of PLMs and their multilingual counterparts (mPLMs) \citep{devlin-etal-2019-bert, conneau-etal-2020-unsupervised} have proven effective for various NLP tasks \citep{artetxe-etal-2020-cross, imanigooghari-etal-2023-glot500}.
However, such models are mostly limited to no more than a hundred languages, as the pre-training requires considerable data that is only available to these languages, leaving the majority of the world's low-resource languages uncovered.
In this work, we explore the possibility of leveraging (1) a PLM in a source language, (2) static word embeddings in a target language, which are readily available for many low-resource languages and are much easier to train, and (3) a technique called model stitching, to enable zero-shot on the target language without the need to pre-train.

Our contribution is summarized as follows: (i) we introduce \textbf{MoSECroT}, 
a novel and challenging task for (especially low-resource) languages where static word embeddings are available. (ii) We propose a solution that leverages relative representations to construct a common space for source (English in our case) and target languages and that allows zero-shot transfer for the target languages.

\section{Related Work}
Aligned crosslingual word embeddings enable transfer learning by benefiting from a shared representation space for the source and target languages. Such embedding pairs are typically either trained jointly \citep{hermann-blunsom-2014-multilingual, DBLP:journals/jair/VulicM16} or obtained through post-alignment \citep{DBLP:conf/iclr/LampleCDR18, artetxe-etal-2018-robust}.
Our work applies a transformation in the manner of the latter to align two embedding spaces where the source embeddings are derived from a PLM and target embeddings are static word embeddings.

Based on a recent consensus that similar inner representations are learned by neural networks regardless of their architecture or domain \citep{pmlr-v97-kornblith19a, vulic-etal-2020-good}, \citet{moschella2022relative} propose an approach to align latent spaces with respect to a set of samples, called parallel anchors. 
They transform the original, absolute space to one defined by relative coordinates of the parallel anchors, and denote all the transformed samples in the relative coordinates as relative representations.

Model stitching was proposed as a way to combine (stitch together) components of different neural models. Trainable stitching layers are first introduced by \citet{DBLP:conf/cvpr/LencV15}, with a series of subsequent works demonstrating the effectiveness of the approach \citep{BianchiSIGIReCom2020, DBLP:conf/nips/BansalNB21}.

\section{MoSECroT Task Setting}
The task setting is straightforward: given a PLM of a high-resource language (regarded as the source language) and static word embeddings of another language (low-resource and regarded as the target language), the goal is to achieve zero-shot transfer by using the target language embeddings directly with the source language model via embedding layer stitching. This can be done by first applying an alignment between the source and target embedding spaces and subsequently swapping the embedding matrices of the PLM.

We propose a novel method that leverages relative representations for embedding space mapping.
In the following, we describe our methodology in more detail.

\section{Methodology} \label{sec:methodology}
\paragraph{Parallel anchor selection}
We first extract bilingual parallel lexica between the source and the target language. For most high-resource languages, large bilingual lexica are available from MUSE\footnote{\url{https://github.com/facebookresearch/MUSE}}.
For low-resource languages, we crawl translations of source language vocabulary from PanLex\footnote{\url{https://panlex.org}} and Google Translate\footnote{\url{https://translate.google.com}}.
Then we derive a subset of the lexica as the parallel anchors $A$ for our method: we only keep those parallel lexica which exist in the embeddings of source and target languages\footnote{The source language is always English and its embeddings are extracted from English BERT's \citep{devlin-etal-2019-bert} token embeddings. For target languages, embeddings are static word embeddings from fastText \citep{bojanowski-etal-2017-enriching}.}.


\paragraph{Relative representations}
Following \citet{moschella2022relative}, we build relative representations (RRs) for each token in the embedding space based on their similarities with anchor tokens in the respective language.
Specifically, we compute the cosine similarity of the embedding of each token with the embedding of each anchor token. 
This computation is done in the embedding spaces of the source and target languages respectively.
For example, in the source language, the similarity between token $x_i$ and anchor $a_j$ is calculated as follows:
$$r^s_{(i, j)} = \text{cos-sim}(\boldsymbol{E}^s_{\{x_i\}}, \boldsymbol{E}^s_{\{a_j\}})$$
where $\boldsymbol{E}^s_{\{x_i\}}$, $\boldsymbol{E}^s_{\{a_j\}}$ are the word embedding of $x_i$ and $a_j$ in the source PLM embeddings $\boldsymbol{E}^s$.
The relative representation of token $x_i$ from the source language is then defined as follows:
$$\boldsymbol{R}^s_{\{x_i\}} = [r^s_{(i, 1)}, r^s_{(i, 2)}, r^s_{(i, 3)}, \cdots, r^s_{(i, |A|)}]$$
Note that the relative representation is sensitive to the order of the anchors, so the relative representation for each token is computed with the anchors in the same order. This computation results in a matrix $\boldsymbol{R}^s \in \mathbb{R}^{|V^s| \times |A|}$ of source language embeddings and a matrix $\boldsymbol{R}^t \in \mathbb{R}^{|V^t| \times |A|}$ of target language embeddings, where $|V^s|$ (resp. $|V^t|$) is the source-language (resp. target-language) vocabulary size and $|A|$ is the number of parallel anchors.

\paragraph{Embedding mapping}
The obtained relative representations are vectors in $\mathbb{R}^{|A|}$ for both source and target languages. This dimension does not suit the hidden dimension of the Transformer body of the source PLM. Therefore, we propose to map the relative representations of both source and target languages back to $\mathbb{R}^{D}$,
which is the same as the dimension of $\boldsymbol{E}^s$.
Given $\boldsymbol{E}^s$ and $\boldsymbol{R}^s$ for source language (resp. $\boldsymbol{E}^t$ and $\boldsymbol{R}^t$ for target language), we compute the transformed embedding of any token $x_i$ from the source language (resp. any token $y_i$ from the target language) as follows:

$$ \boldsymbol{F}^{s}_{\{x_i\}} = \frac{\sum_{n \in \mathbb{N}(x_i)} ( \boldsymbol{R}^s_{\{x_i\}, n} / \tau \cdot \boldsymbol{E}^{s}_{\{n\}})}{\sum_{n \in \mathbb{N}(x_i)} \boldsymbol{R}^s_{\{x_i\}, n} / \tau}$$
$$ \boldsymbol{F}^{t}_{\{y_i\}} = \frac{\sum_{n \in \mathbb{N}(y_i)} (\boldsymbol{R}^t_{\{y_i\}, n} / \tau \cdot \boldsymbol{E}^{s}_{\{n\}})}{\sum_{n \in \mathbb{N}(y_i)} \boldsymbol{R}^t_{\{y_i\}, n} / \tau}$$
where $\mathbb{N}(x_i)$ (resp. $\mathbb{N}(y_i)$) is the set of top-$k$ closest anchors in terms of the cosine similarity recorded in $\boldsymbol{R}^s_{x_i}$ (resp. $\boldsymbol{R}^t_{y_i}$), $\boldsymbol{R}^s_{\{x_i\}, n}$ (resp. $\boldsymbol{R}^s_{\{y_i\}, n}$) is the cosine similarity between $\boldsymbol{E}^s_{\{x_i\}}$ (resp. $\boldsymbol{E}^t_{\{y_i\}}$) and $\boldsymbol{E}^s_{\{n\}}$ (resp. $\boldsymbol{E}^t_{\{n\}}$), and $\tau$ is the temperature. Note that both the resulting transformed embeddings $\boldsymbol{F}^{s}_{\{x_i\}}$ and $\boldsymbol{F}^{t}_{\{y_i\}}$ are in $\mathbb{R}^D$, because it is a weighted sum of the anchor embedding in the \textbf{source language}, i.e., $\boldsymbol{E}^{s}_{\{n\}}$. A simple summary of the process is to \emph{represent any token, no matter whether it is from the source or target language, as a weighted sum of the embeddings of some parallel anchors in the source-language embedding space.}

\paragraph{Zero-shot stitching}
So far we project the target-language embeddings to $\mathbb{R}^D$,
which suits the hidden dimension of the Transformer body of the source language.
We also manipulate the original token embedding matrix of the source language, where the matrix dimensions stay the same: $\boldsymbol{F}^s \in \mathbb{R}^{|V^s| \times D}$. 
We can simply fine-tune the model ($\boldsymbol{F}^s$ and the Transformer body) on the source-language train set of a downstream task and then assemble a target-language model for zero-shot transfer, without training on the target language. 
To do this, we only need to swap the source-language embeddings $\boldsymbol{F}^s$ with target-language embeddings $\boldsymbol{F}^t$.

\section{Experiments}
\subsection{Setup}
We use the cased version of the English BERT model (\texttt{bert-base-cased}) as the source language PLM and consider eight target languages.
Three of the target languages are high-resource: German (\textbf{de}), Spanish (\textbf{es}), and Chinese (\textbf{zh}),
and the rest are low-resource: Faroese (\textbf{fo}), Maltese (\textbf{mt}), Eastern Low German (\textbf{nds}), Sakha (\textbf{sah}), and Tatar (\textbf{tt}).
Pre-trained static embeddings for all target languages are available
from fastText\footnote{\url{https://fasttext.cc/docs/en/pretrained-vectors.html}}, except for
Eastern Low German, for which we download fastText embeddings from Huggingface\footnote{\url{https://huggingface.co/facebook/fasttext-nds-vectors}}.

Using the method proposed in \secref{methodology}, we obtain pairwise parallel anchors between English and each target language.
The size of the anchor set varies depending on the vocabulary size of the language's embeddings and the overlap between the English and target language lexica, which is the following for each target language: 11836 (en-de), 11395 (en-es), 7662 (en-zh), 1577 (en-fo), 2600 (en-mt), 1309 (nds), 3242 (en-sah), and 9275 (en-tt).

We evaluate the proposed method on two text classification datasets: Multilingual Amazon Reviews Corpus \citep{keung-etal-2020-multilingual} and Taxi1500 \citep{ma2023taxi1500}. See \secref{eval_datasets} for details.





\seclabel{rr_settings}
Apart from the standard weighting scheme illustrated in \secref{methodology}, we propose two more settings: one where we apply softmax over relative representation weights (in the \textbf{Embedding mapping} step), and another using sparsemax \citep{pmlr-v48-martins16}. 
Compared to softmax, sparsemax produces sparse weight distributions, meaning more similarities are concentrated on fewer anchors.
We conduct preliminary experiments to identify the optimal top-$k$ closest anchors $\in \{1, 10, 50, 100\}$ and find that the results are best when using the top 50 anchors. See \secref{num_anchors} for an exploration of how different choices of $k$ influence the performance.

\begin{table}
    \centering
    \begin{tabular}{l|ccc}
         &  de&  es& zh \\
         \toprule
         LR&  0.52& 0.51& 0.50 \\
         mBERT&  \textbf{0.61}&  \textbf{0.65}& \textbf{0.51} \\
         LS&  0.46&  0.46& 0.30 \\ \midrule
         RRs standard top-50&  0.53&  0.51& 0.38 \\
         RRs softmax top-50&  0.50&  0.53& 0.38\\
         RRs sparsemax top-50&  0.56&  0.57& 0.24 \\
    \end{tabular}
    \caption{Evaluation results on the Amazon Reviews Corpus. We report macro $F_1$ scores on the test sets of three high-resource target languages.
    \textbf{Bold}: highest score per column.}
    \label{tab:amazon_reviews}
\end{table}

\begin{table*}[t]
    \centering
    \begin{tabular}{l|cccccccc}
         &  de&  es&  zh&  mt&  sah&  fo&  nds& tt\\
         \toprule
         LR& \textbf{ 0.30}&  0.32&  0.56&  \textbf{0.38}&  \textbf{0.48}&  \textbf{0.47}&  \textbf{0.18}& \textbf{0.43}\\
         mBERT&  0.24&  \textbf{0.60}&  \textbf{0.62}&  0.08&  0.07&  0.18&  0.12& 0.18\\
         LS&  0.14&  0.26&  0.24&  0.08&  0.12&  0.06&  0.08& 0.07\\
         \midrule
         RRs standard top50&  0.20&  0.44&  0.28&  0.14&  0.16&  0.16&  0.06& 0.14\\
         RRs softmax top50&  0.20&  0.48&  0.28&  0.15&  0.19&  0.16&  0.06& 0.17\\
         RRs sparsemax top50&  0.24&  0.37&  0.13&  0.15&  0.18&  0.20&  0.13& 0.21\\
    \end{tabular}
    \caption{Evaluation results on the Taxi1500 dataset. Reported metrics are macro $F_1$ scores on the test sets of eight target languages.
    Scores are averaged over five runs with different random seeds.
    \textbf{Bold}: highest score per column.}
    \label{tab:taxi1500}
\end{table*}

\subsection{Baselines}
We compare our method against three baselines:

\paragraph{Logistic Regression (LR)}
We train a simple target language logistic regression classifier using the average of static word embeddings of the input sentences. This approach does not require expensive training of a language model but assumes we have sufficient target language training data for a specific downstream task, which is hardly the case for most low-resource languages in real scenarios.

\paragraph{mBERT}
We fine-tune multilingual BERT (mBERT) \citep{devlin-etal-2019-bert}, 
which is pre-trained on more than 100 languages,
using the English training data, and perform zero-shot predictions directly on the target language test data.

\paragraph{Least squares projection (LS)}
We propose a straightforward approach, inspired by embedding alignment frameworks such as VecMap \citep{artetxe-etal-2018-robust}, to project target language embeddings into the same space as the English PLM embeddings.
Specifically, we learn a transformation matrix $\boldsymbol{W} \in \mathbb{R}^{D^t \times D}$ by minimizing $||\boldsymbol{A}^t \boldsymbol{W} - \boldsymbol{A}^s ||_F^2$, where $\boldsymbol{A}^t \in \mathbb{R}^{|A| \times D^t}$ is the embeddings of anchors in the target language and $\boldsymbol{A}^s \in \mathbb{R}^{|A| \times D}$ is the embeddings of anchors from the English PLM.
We then project all target language embeddings using $\boldsymbol{W}$ and replace the BERT embedding layer with the resulting matrix.


\subsection{Results}
We present evaluation results of RRs with the proposed settings (\secref{rr_settings}) and compare them with the baselines in Tables \ref{tab:amazon_reviews} and \ref{tab:taxi1500}. Macro $F_1$ is used due to class imbalance in both datasets.

We notice that the naive LS baseline is almost always beaten by the proposed method under multiple RR settings on both datasets. The only exception is nds, in Table \ref{tab:taxi1500}, where both LS and RRs perform badly. This observation is a strong indicator that RRs can better leverage the semantic similarity encoded in different types of embeddings than LS.

Not very surprisingly, zero-shot with mBERT is effective for high-resource languages in both datasets but underperforms LR with large gaps on low-resource languages in Taxi1500. There are two possible explanations for this phenomenon. First, representations in mBERT are not well-aligned across low-resource languages. This is possibly due to data sparsity, which is observed by previous work \citep{wu-dredze-2020-languages}, where mBERT archives good performance on high-resource languages but sub-optimal performance on low-resource languages.
Second, Taxi1500 is a relatively easy task: a model with good alignment across languages, especially on the word level, is expected to perform well.
This argument is supported by a previous work \citep{liu-etal-2023-crosslingual-transfer}, where well-aligned word embeddings achieve better zero-shot crosslingual performance than mPLMs on a wide range of languages in Taxi1500.

Although none of the RR settings outperforms mBERT on high-resource languages (as mentioned earlier, mBERT has strong crosslingual transfer ability on high-resource languages), for all five low-resource languages not seen by mBERT (mt, sah, fo, nds, tt), RRs outperform mBERT consistently, with varying margins (ranging from +0.12 for sah to +0.01 for nds).
This suggests that RRs can be a promising alternative when a low-resource language is not covered by an mPLM.

\section{Analysis}
In this section, we want to propose possible reasons for the suboptimal results obtained by our framework tackling the MoSECroT task.

\paragraph{Anchor selection}
The quality of the parallel anchors largely relies on the quality of the bilingual lexica, which may contain, among others, polysemous words, that may influence the alignment quality.
Normalization can also be a source of ambiguity.
For example, MUSE converts all words into lowercase, so the word \texttt{sie} can have three meanings in the German-English lexicon: \texttt{you}, \texttt{she}, and \texttt{they}.
We (1) only consider one translation (if there are multiple)
for each target language word, which may not be the most accurate one; and
(2) treat all target language words whose translations are in the source language vocabulary as anchors, which increases the frequency of noisy translation pairs.

We try to decrease the influence of potentially noisy anchor pairs by reducing the number of anchors to 3000 and 500 (the original anchor set used during the preliminary experiments contains 6731 anchors, see \secref{methodology})
through random sampling, following the observation by \citet{moschella2022relative} that uniform selection from an anchor set is both straightforward and has good performance.
We also remove stop words, whose translations are more unstable, from the anchor set. Neither of the two modifications shows an improvement over the full anchor set (see \secref{total_number} for the comparison).
One possible explanation is that the translation qualities vary across anchors and thus we cannot predict the quality of sampled anchors.

\paragraph{Translation quality}
We find that a large portion of translations retrieved from PanLex are of low quality. This is partly due to PanLex using intermediate languages when direct translation is unavailable for the language pair.
We filter the translations by empirically setting a threshold to the translation quality scores, available through the API for every translation.
Nevertheless, we note that a high translation quality score does not guarantee the translation is perfect, and many translations are good despite having low translation quality scores.
We believe the lack of high-quality parallel lexica is a possible reason that RRs do not reach their full potential on low-resource languages.

\paragraph{Reinitialized embedding space}
Our method requires swapping the original PLM embeddings with the transformed English RRs before fine-tuning on English data, whereas the embedding space of RRs might diverge substantially from the original embedding space.
As a result, it is unclear whether the rest of the model parameters can be adapted to the new embeddings during fine-tuning, especially on smaller datasets like Taxi1500.
We thus suggest the alteration of the embedding space through reinitialization with RRs as a likely factor as to why we do not achieve good performance.

\section{Conclusion}
In this work, we introduce MoSECroT, a novel and challenging task that is relevant for, in particular, low-resource languages for which static word embeddings are available but few resources exist.
In addition, we propose for the first time a method that leverages relative representations for embedding space mapping and enables zero-shot transfer.
Specifically, we fine-tune a monolingual English language model using only English data, swap the embeddings with target language embeddings aligned using RRs, and apply zero-shot evaluation on the target language.
We show that the proposed method is promising compared with mBERT on unseen languages but only modest improvements are achieved.
We provide several possible reasons and leave improvement possibilities for future research.

\section*{Limitations}
In this work, we propose the task of MoSECroT and a solution to leverage available static pre-trained embeddings and tackle downstream tasks for low-resource languages.
Our work has a few limitations open to future research.
First, we only experiment with one model architecture (BERT). Although many language-specific BERT models exist and thus our method is applicable to a wide range of high-resource source languages, it would nevertheless be interesting to compare performance across different model architectures.
Second, the explored tasks are exclusively text classification tasks. We expect that the robustness of our method can be much better studied by applying it to a more diverse set of tasks.

\section*{Acknowledgements}
This work was funded by the European Research Council (grant \#740516).

\bibliography{anthology,custom}

\begin{thebibliography}{20}
\expandafter\ifx\csname natexlab\endcsname\relax\def\natexlab#1{#1}\fi

\bibitem[{Artetxe et~al.(2018)Artetxe, Labaka, and Agirre}]{artetxe-etal-2018-robust}
Mikel Artetxe, Gorka Labaka, and Eneko Agirre. 2018.
\newblock \href {https://doi.org/10.18653/v1/P18-1073} {A robust self-learning method for fully unsupervised cross-lingual mappings of word embeddings}.
\newblock In \emph{Proceedings of the 56th Annual Meeting of the Association for Computational Linguistics (Volume 1: Long Papers)}, pages 789--798, Melbourne, Australia. Association for Computational Linguistics.

\bibitem[{Artetxe et~al.(2020)Artetxe, Ruder, and Yogatama}]{artetxe-etal-2020-cross}
Mikel Artetxe, Sebastian Ruder, and Dani Yogatama. 2020.
\newblock \href {https://doi.org/10.18653/v1/2020.acl-main.421} {On the cross-lingual transferability of monolingual representations}.
\newblock In \emph{Proceedings of the 58th Annual Meeting of the Association for Computational Linguistics}, pages 4623--4637, Online. Association for Computational Linguistics.

\bibitem[{Bansal et~al.(2021)Bansal, Nakkiran, and Barak}]{DBLP:conf/nips/BansalNB21}
Yamini Bansal, Preetum Nakkiran, and Boaz Barak. 2021.
\newblock \href {https://proceedings.neurips.cc/paper/2021/hash/01ded4259d101feb739b06c399e9cd9c-Abstract.html} {Revisiting model stitching to compare neural representations}.
\newblock In \emph{Advances in Neural Information Processing Systems 34: Annual Conference on Neural Information Processing Systems 2021, NeurIPS 2021, December 6-14, 2021, virtual}, pages 225--236.

\bibitem[{Bianchi et~al.(2020)Bianchi, Tagliabue, Yu, Bigon, and Greco}]{BianchiSIGIReCom2020}
Federico Bianchi, Jacopo Tagliabue, Bingqing Yu, Luca Bigon, and Ciro Greco. 2020.
\newblock \href {https://arxiv.org/abs/2007.14906} {Fantastic embeddings and how to align them: Zero-shot inference in a multi-shop scenario}.
\newblock In \emph{Proceedings of the SIGIR 2020 eCom workshop, July 2020, Virtual Event, published at http://ceur-ws.org (to appear)}.

\bibitem[{Bojanowski et~al.(2017)Bojanowski, Grave, Joulin, and Mikolov}]{bojanowski-etal-2017-enriching}
Piotr Bojanowski, Edouard Grave, Armand Joulin, and Tomas Mikolov. 2017.
\newblock \href {https://doi.org/10.1162/tacl_a_00051} {Enriching word vectors with subword information}.
\newblock \emph{Transactions of the Association for Computational Linguistics}, 5:135--146.

\bibitem[{Conneau et~al.(2020)Conneau, Khandelwal, Goyal, Chaudhary, Wenzek, Guzm{\'a}n, Grave, Ott, Zettlemoyer, and Stoyanov}]{conneau-etal-2020-unsupervised}
Alexis Conneau, Kartikay Khandelwal, Naman Goyal, Vishrav Chaudhary, Guillaume Wenzek, Francisco Guzm{\'a}n, Edouard Grave, Myle Ott, Luke Zettlemoyer, and Veselin Stoyanov. 2020.
\newblock \href {https://doi.org/10.18653/v1/2020.acl-main.747} {Unsupervised cross-lingual representation learning at scale}.
\newblock In \emph{Proceedings of the 58th Annual Meeting of the Association for Computational Linguistics}, pages 8440--8451, Online. Association for Computational Linguistics.

\bibitem[{Devlin et~al.(2019)Devlin, Chang, Lee, and Toutanova}]{devlin-etal-2019-bert}
Jacob Devlin, Ming-Wei Chang, Kenton Lee, and Kristina Toutanova. 2019.
\newblock \href {https://doi.org/10.18653/v1/N19-1423} {{BERT}: Pre-training of deep bidirectional transformers for language understanding}.
\newblock In \emph{Proceedings of the 2019 Conference of the North {A}merican Chapter of the Association for Computational Linguistics: Human Language Technologies, Volume 1 (Long and Short Papers)}, pages 4171--4186, Minneapolis, Minnesota. Association for Computational Linguistics.

\bibitem[{Hermann and Blunsom(2014)}]{hermann-blunsom-2014-multilingual}
Karl~Moritz Hermann and Phil Blunsom. 2014.
\newblock \href {https://doi.org/10.3115/v1/P14-1006} {Multilingual models for compositional distributed semantics}.
\newblock In \emph{Proceedings of the 52nd Annual Meeting of the Association for Computational Linguistics (Volume 1: Long Papers)}, pages 58--68, Baltimore, Maryland. Association for Computational Linguistics.

\bibitem[{ImaniGooghari et~al.(2023)ImaniGooghari, Lin, Kargaran, Severini, Jalili~Sabet, Kassner, Ma, Schmid, Martins, Yvon, and Sch{\"u}tze}]{imanigooghari-etal-2023-glot500}
Ayyoob ImaniGooghari, Peiqin Lin, Amir~Hossein Kargaran, Silvia Severini, Masoud Jalili~Sabet, Nora Kassner, Chunlan Ma, Helmut Schmid, Andr{\'e} Martins, Fran{\c{c}}ois Yvon, and Hinrich Sch{\"u}tze. 2023.
\newblock \href {https://doi.org/10.18653/v1/2023.acl-long.61} {Glot500: Scaling multilingual corpora and language models to 500 languages}.
\newblock In \emph{Proceedings of the 61st Annual Meeting of the Association for Computational Linguistics (Volume 1: Long Papers)}, pages 1082--1117, Toronto, Canada. Association for Computational Linguistics.

\bibitem[{Keung et~al.(2020)Keung, Lu, Szarvas, and Smith}]{keung-etal-2020-multilingual}
Phillip Keung, Yichao Lu, Gy{\"o}rgy Szarvas, and Noah~A. Smith. 2020.
\newblock \href {https://doi.org/10.18653/v1/2020.emnlp-main.369} {The multilingual {A}mazon reviews corpus}.
\newblock In \emph{Proceedings of the 2020 Conference on Empirical Methods in Natural Language Processing (EMNLP)}, pages 4563--4568, Online. Association for Computational Linguistics.

\bibitem[{Kornblith et~al.(2019)Kornblith, Norouzi, Lee, and Hinton}]{pmlr-v97-kornblith19a}
Simon Kornblith, Mohammad Norouzi, Honglak Lee, and Geoffrey Hinton. 2019.
\newblock \href {https://proceedings.mlr.press/v97/kornblith19a.html} {Similarity of neural network representations revisited}.
\newblock In \emph{Proceedings of the 36th International Conference on Machine Learning}, volume~97 of \emph{Proceedings of Machine Learning Research}, pages 3519--3529. PMLR.

\bibitem[{Lample et~al.(2018)Lample, Conneau, Denoyer, and Ranzato}]{DBLP:conf/iclr/LampleCDR18}
Guillaume Lample, Alexis Conneau, Ludovic Denoyer, and Marc'Aurelio Ranzato. 2018.
\newblock \href {https://openreview.net/forum?id=rkYTTf-AZ} {Unsupervised machine translation using monolingual corpora only}.
\newblock In \emph{6th International Conference on Learning Representations, {ICLR} 2018, Vancouver, BC, Canada, April 30 - May 3, 2018, Conference Track Proceedings}. OpenReview.net.

\bibitem[{Lenc and Vedaldi(2015)}]{DBLP:conf/cvpr/LencV15}
Karel Lenc and Andrea Vedaldi. 2015.
\newblock \href {https://doi.org/10.1109/CVPR.2015.7298701} {Understanding image representations by measuring their equivariance and equivalence}.
\newblock In \emph{{IEEE} Conference on Computer Vision and Pattern Recognition, {CVPR} 2015, Boston, MA, USA, June 7-12, 2015}, pages 991--999. {IEEE} Computer Society.

\bibitem[{Liu et~al.(2023)Liu, Ye, Weissweiler, Pei, and Schuetze}]{liu-etal-2023-crosslingual-transfer}
Yihong Liu, Haotian Ye, Leonie Weissweiler, Renhao Pei, and Hinrich Schuetze. 2023.
\newblock \href {https://doi.org/10.18653/v1/2023.findings-emnlp.562} {Crosslingual transfer learning for low-resource languages based on multilingual colexification graphs}.
\newblock In \emph{Findings of the Association for Computational Linguistics: EMNLP 2023}, pages 8376--8401, Singapore. Association for Computational Linguistics.

\bibitem[{Ma et~al.(2023)Ma, ImaniGooghari, Ye, Asgari, and Schütze}]{ma2023taxi1500}
Chunlan Ma, Ayyoob ImaniGooghari, Haotian Ye, Ehsaneddin Asgari, and Hinrich Schütze. 2023.
\newblock \href {http://arxiv.org/abs/2305.08487} {Taxi1500: A multilingual dataset for text classification in 1500 languages}.

\bibitem[{Martins and Astudillo(2016)}]{pmlr-v48-martins16}
Andre Martins and Ramon Astudillo. 2016.
\newblock \href {https://proceedings.mlr.press/v48/martins16.html} {From softmax to sparsemax: A sparse model of attention and multi-label classification}.
\newblock In \emph{Proceedings of The 33rd International Conference on Machine Learning}, volume~48 of \emph{Proceedings of Machine Learning Research}, pages 1614--1623, New York, New York, USA. PMLR.

\bibitem[{Moschella et~al.(2023)Moschella, Maiorca, Fumero, Norelli, Locatello, and Rodol{\`{a}}}]{moschella2022relative}
Luca Moschella, Valentino Maiorca, Marco Fumero, Antonio Norelli, Francesco Locatello, and Emanuele Rodol{\`{a}}. 2023.
\newblock \href {https://openreview.net/pdf?id=SrC-nwieGJ} {Relative representations enable zero-shot latent space communication}.
\newblock In \emph{The Eleventh International Conference on Learning Representations, {ICLR} 2023, Kigali, Rwanda, May 1-5, 2023}. OpenReview.net.

\bibitem[{Vulic and Moens(2016)}]{DBLP:journals/jair/VulicM16}
Ivan Vulic and Marie{-}Francine Moens. 2016.
\newblock \href {https://doi.org/10.1613/JAIR.4986} {Bilingual distributed word representations from document-aligned comparable data}.
\newblock \emph{J. Artif. Intell. Res.}, 55:953--994.

\bibitem[{Vuli{\'c} et~al.(2020)Vuli{\'c}, Ruder, and S{\o}gaard}]{vulic-etal-2020-good}
Ivan Vuli{\'c}, Sebastian Ruder, and Anders S{\o}gaard. 2020.
\newblock \href {https://doi.org/10.18653/v1/2020.emnlp-main.257} {Are all good word vector spaces isomorphic?}
\newblock In \emph{Proceedings of the 2020 Conference on Empirical Methods in Natural Language Processing (EMNLP)}, pages 3178--3192, Online. Association for Computational Linguistics.

\bibitem[{Wu and Dredze(2020)}]{wu-dredze-2020-languages}
Shijie Wu and Mark Dredze. 2020.
\newblock \href {https://doi.org/10.18653/v1/2020.repl4nlp-1.16} {Are all languages created equal in multilingual {BERT}?}
\newblock In \emph{Proceedings of the 5th Workshop on Representation Learning for NLP}, pages 120--130, Online. Association for Computational Linguistics.

\end{thebibliography}

\appendix

\section{Number of closest anchors}\seclabel{num_anchors}
In addition to using all (6731) parallel anchors, we consider only the top-$k$ ($k \in \{1, 10, 50, 100\}$) closest anchors of each word.
We identify the optimal value for $k$ closest anchors based on zero-shot performance on German and Chinese portions of the Amazon Reviews Corpus (\secref{amazon}).
Table \ref{tab:k_anchors} shows results for different $k$ values.

\begin{table}[h!]
    \centering
    \begin{tabular}{r|cc}
         $k$ &  de & zh \\
         \toprule
         1 &  0.44 &  0.41 \\
         10&  0.51 &  0.38 \\
         50&  0.50 &  0.40 \\
         100&  0.51 &  0.38 \\
         6731& 0.44 & 0.21 \\
    \end{tabular}
    \caption{Number of closest parallel anchors ($k$) and the corresponding zero-shot performance on de and zh portions of the Amazon Reviews Corpus.}
    \label{tab:k_anchors}
\end{table}

\section{Total number of anchors}\seclabel{total_number}
Following \citet{moschella2022relative}, we randomly sample a subset of the parallel anchors ($|\boldsymbol{A}| \in \{500, 3000\}$), and exclude stop words from the anchor set. Table \ref{tab:total_anchors} shows zero-shot performance on German and Chinese portions of the Amazon Reviews Corpus (\secref{amazon}).

\begin{table}[h!]
    \centering
    \begin{tabular}{r|cc}
         $|\boldsymbol{A}|$ &  de & zh \\
         \toprule
         500 &  0.39 &  0.19 \\
         3000&  0.19 &  0.19 \\
         6731&  0.44 &  0.21 \\
    \end{tabular}
    \caption{The total number of parallel anchors and the corresponding zero-shot performance on de and zh portions of the Amazon Reviews Corpus.}
    \label{tab:total_anchors}
\end{table}

\section{Evaluation datasets} \seclabel{eval_datasets}
\subsection{Multilingual Amazon Reviews Corpus} \seclabel{amazon}
Presented by \citet{keung-etal-2020-multilingual} and containing product reviews in six languages, the original dataset uses five labels corresponding to star ratings, which we aggregate into three classes: positive, neutral, and negative.
We evaluate the three high-resource target languages (de, es, zh) on this dataset.

\subsection{Taxi1500}
Taxi1500 \citep{ma2023taxi1500} is a classification dataset containing six classes for more than 1500 languages, including all of our target languages.
We follow the authors' original training procedure and hyperparameters and use a learning rate of 1e-5 instead of 2e-5, which we find works better for our settings.

\section{Computational resources}
Training can be completed in under three hours on eight NVIDIA GeForce GTX 1080 Ti GPUs for the Multilingual Amazon Reviews Corpus or about half an hour on a single NVIDIA GeForce GTX 1080 Ti GPU for Taxi1500.

\end{document}